\documentclass[runningheads]{llncs}

\usepackage{booktabs}
\usepackage{placeins}
\usepackage{amssymb}

\usepackage{graphicx}
\begin{document}
\title{ViLegalExpert: A Large-Scale Benchmark for Vietnamese Legal Retrieval and Question Answering from Real-World Consultations}
%
%
\author{Dat Tien Nguyen\inst{1,2} \and Nghia Hieu Nguyen\inst{1,2} \and Anh Thi-Hoang Nguyen\inst{1,2} \and Dung Ha Nguyen\inst{1,2} \and Kiet Van Nguyen\inst{1,2} \and Ngan Luu-Thuy Nguyen\inst{1,2}}
\authorrunning{Nguyen et al.}
%
\institute{University of Information Technology \and
Vietnam National University, Ho Chi Minh city, Viet Nam \\
\email{23520262@gm.uit.edu.vn, \{nghiangh, anhnth, dungnh, kietnv, ngannlt\}@uit.edu.vn}
}
\maketitle              
\begin{abstract}
Trustworthy Legal AI requires systems that can answer legal questions while grounding their responses in authoritative sources. However, existing Vietnamese legal benchmarks provide limited coverage of real-world legal consultations. We introduce \textbf{ViLegalExpert}, a large-scale benchmark constructed from authentic citizen--lawyer consultations, containing over \textbf{172K} questions across \textbf{34 legal domains}, together with professional answers and expert-verified legal evidence. ViLegalExpert supports legal information retrieval, extractive QA, and abstractive QA. Experiments with representative retrieval methods and language models reveal substantial challenges in evidence retrieval and grounded answer generation. While pretrained models perform strongly on QA, hybrid retrieval achieves the best retrieval performance. These results demonstrate the difficulty of mapping naturally expressed legal questions to authoritative provisions and establish ViLegalExpert as a challenging benchmark for reliable Vietnamese Legal AI.
\end{abstract}

\section{Introduction}
\label{sec:introduction}

Recent advances in Large Language Models (LLMs) have accelerated the development of Legal Artificial Intelligence (Legal AI), enabling applications such as legal assistants and automated legal consultation \cite{zhong-etal-2020-nlp}. However, legal applications require not only fluent responses but also accurate answers grounded in authoritative legal sources \cite{inbook,vold2021using,khazaeli-etal-2021-free,trautmann-etal-2024-measuring}. This requirement has become particularly important with Retrieval-Augmented Generation (RAG), where retrieval quality directly affects the reliability of generated answers \cite{DBLP:journals/corr/abs-2005-11401}. Consequently, realistic benchmarks that jointly evaluate legal retrieval and question answering are essential for developing trustworthy Legal AI systems.

Several Vietnamese legal datasets have recently been introduced for information retrieval, question answering, and legal reasoning. Despite their contributions, existing resources are largely constructed for curated or task-specific settings, including shared tasks, manually annotated questions, and synthetic evaluation scenarios. They therefore provide limited coverage of the diverse information needs and linguistic characteristics of real-world legal consultations. Moreover, the limited scale or domain coverage of existing question-based resources poses challenges for evaluating modern retrieval and LLM-based systems.

To address these limitations, we introduce \textbf{ViLegalExpert}, a large-scale Vietnamese benchmark constructed from authentic online consultations between citizens and legal professionals. ViLegalExpert contains over \textbf{172K} real-world legal questions spanning \textbf{34 legal domains}, together with professional answers and expert-verified provision-level legal evidence. These annotations enable a unified evaluation of \textbf{legal information retrieval, extractive QA, and abstractive QA}, providing a realistic setting for assessing both evidence retrieval and answer generation.

We benchmark representative retrieval methods, pretrained language models, and instruction-tuned LLMs on ViLegalExpert. Our experiments reveal substantial challenges in mapping naturally expressed legal questions to authoritative provisions and generating correctly grounded answers, highlighting important limitations of current approaches.

Our main contributions are:
\begin{itemize}
    \item We introduce \textbf{ViLegalExpert}, a large-scale Vietnamese legal benchmark containing over 172K authentic consultation questions across 34 legal domains, together with professional answers and expert-verified legal evidence.
    \item We establish a unified benchmark for \textbf{legal retrieval, extractive QA, and abstractive QA}, enabling evaluation of both evidence retrieval and answer generation.
    \item We provide comprehensive baselines and analyses across representative retrieval and language models, revealing key challenges for reliable and evidence-grounded Vietnamese Legal AI.
\end{itemize}

\section{Related Work}

\begin{table*}[t]
    \centering
    \caption{Comparison of representative legal retrieval and question answering benchmarks. ViLegalExpert is the largest Vietnamese benchmark constructed from authentic citizen--lawyer consultation records and supports both legal information retrieval and question answering across 34 legal domains.}
    \label{tab:dataset_comparison}
    \resizebox{\linewidth}{!}{
    \begin{tabular}{llcccccc}
    \toprule
    \textbf{Dataset} &
    \textbf{Lang.} &
    \textbf{Year} &
    \textbf{Source} &
    \textbf{Retrieval} &
    \textbf{QA} &
    \textbf{\#Samples} &
    \textbf{\#Domains} \\
    \midrule
    
    COLIEE                & EN & 2014 & Bar exam / Case law              & \checkmark & \checkmark & --      & 1 \\
    JEC-QA                & ZH & 2020   & Judicial examination             & $\times$   & \checkmark & 28,641  & 13 \\
    ALQAC                 & JA & 2020   & Legal documents                  & \checkmark & \checkmark & 272     & 4 \\
    EQUALS                & EN & 2023   & Legal documents                  & $\times$   & \checkmark & 8,448   & -- \\
    \midrule
    
    Zalo AI Challenge     & VI & 2021   & Competition                      & \checkmark & $\times$   & --      & -- \\
    ALQAC 2022            & VI & 2022   & Legal documents                  & \checkmark & \checkmark & 520     & 4 \\
    ALQAC 2023            & VI & 2023   & Legal documents                  & \checkmark & \checkmark & 1,200+  & 4 \\
    SoICT LegalIR         & VI & 2024   & Competition                      & \checkmark & $\times$   & 10,000  & -- \\
    ViRHE4QA              & VI & 2024   & Curated QA                       & \checkmark & \checkmark & 9,758   & -- \\
    VLSP LegalIR          & VI & 2024   & Competition                      & \checkmark & $\times$   & 2,000   & -- \\
    VLegal-Bench          & VI & 2025   & Curated benchmark                & \checkmark & \checkmark & 10,450  & -- \\
    VLQA                  & VI & 2025   & Annotated legal QA               & \checkmark & \checkmark & 24,012  & 27 \\
    \midrule
    
    \textbf{ViLegalExpert (Ours)}
    & \textbf{VI}
    & \textbf{2026}
    & \textbf{Citizen--lawyer consultations}
    & \checkmark
    & \checkmark
    & \textbf{177,158}
    & \textbf{34} \\
    
    \bottomrule
    \end{tabular}}
\end{table*}

\subsection{Legal NLP Benchmarks}

Benchmark datasets have played a fundamental role in advancing Legal Natural Language Processing (Legal NLP) by providing standardized evaluation protocols for a wide range of legal tasks. Existing benchmarks cover legal document classification \cite{chalkidis-etal-2019-large,tuggener-etal-2020-ledgar}, legal information retrieval \cite{10.1145/3769126.3785016,louis-spanakis-2022-statutory}, legal summarization \cite{10.1016/j.jksuci.2019.11.015,10.1007/s10462-017-9566-2}, legal entailment, and legal question answering \cite{zhong2019jecqalegaldomainquestionanswering,inproceedings}. More recently, the emergence of Retrieval-Augmented Generation (RAG) and Large Language Models (LLMs) has further increased the importance of benchmarks that jointly evaluate document retrieval and evidence-grounded answer generation. These resources have substantially accelerated the development of Legal AI in high-resource languages such as English and Chinese by enabling consistent comparison of retrieval models, pretrained language models, and LLM-based systems.

\subsection{Vietnamese Legal NLP Resources}

Vietnamese Legal NLP has witnessed rapid progress in recent years, driven by the release of several benchmark datasets and shared tasks. Early efforts focused on legal text retrieval and question answering through the ALQAC shared tasks, which introduced benchmark datasets for legal retrieval and answer extraction. Subsequent work expanded to larger retrieval corpora and more realistic retrieval settings, including the SoICT LegalIR Hackathon, ViRHE4QA, and the VLSP LegalIR shared task. Recent benchmarks have further broadened the scope of Vietnamese Legal AI by introducing datasets for legal natural language inference (ViLegalNLI), legal reasoning and Retrieval-Augmented Generation (VLegal-Bench), and multilingual legal question answering (VLQA) \cite{11063484,nguyen2025vlqacomprehensivelargehighquality}.

Despite these advances, existing Vietnamese resources exhibit several limitations. First, many benchmarks are designed for specific shared tasks or individual downstream problems, such as retrieval, question answering, or natural language inference, making it difficult to evaluate Legal AI systems under a unified framework. Second, many datasets rely on manually curated evaluation sets or questions constructed from legal documents, which only partially reflect the diverse information needs encountered in real legal consultation. Finally, although some benchmarks provide large legal corpora, the number of authentic legal questions available for training and evaluation remains relatively limited.

\subsection{Comparison with Existing Benchmarks}

Table~\ref{tab:dataset_comparison} compares ViLegalExpert with representative Vietnamese and multilingual legal benchmarks. Unlike previous datasets that primarily focus on individual tasks or curated benchmark settings, ViLegalExpert is constructed from authentic legal consultation forums, where citizens seek legal advice from legal professionals. Consequently, the dataset captures naturally occurring legal information needs expressed in everyday language, providing a more realistic evaluation setting for Legal AI.

ViLegalExpert contains over 177K authentic legal questions spanning 34 legal domains, together with corresponding answers and a structured retrieval corpus supporting both legal information retrieval and question answering. Compared with existing Vietnamese benchmarks, ViLegalExpert substantially expands the scale of authentic legal consultation questions while unifying retrieval and question answering under a single benchmark. We believe these characteristics make ViLegalExpert a valuable resource for developing, benchmarking, and analyzing modern Vietnamese Legal AI systems, particularly retrieval-based and LLM-based approaches.

\section{The ViLegalExpert Dataset}

\subsection{Dataset Construction} \label{sec:dataset_construction}

The construction of ViLegalExpert consists of four main stages: (1) collecting real-world legal questions and answers, (2) constructing a corpus of authoritative Vietnamese legal documents, (3) linking questions to relevant legal documents through expert annotation, and (4) formatting and splitting the resulting data.

\subsubsection{Collection of Legal Questions and Answers.}
We first collect question--answer pairs from publicly accessible online legal consultation platforms, where citizens can submit questions about legal issues and receive answers from legal professionals. These questions naturally arise from practical situations encountered in everyday life, rather than being manually written for benchmark construction. As a result, they exhibit substantial diversity in writing style, level of detail, and legal information needs. Each collected instance consists of a citizen's question and its corresponding answer provided by a legal professional. The collected questions cover a broad range of legal topics, including civil, criminal, labor, land, marriage and family, business, and administrative law.

\subsubsection{Collection of Legal Documents.}
In parallel, we construct a legal document corpus from official Vietnamese government sources. The corpus contains different types of normative legal documents, including the Constitution, laws, decrees, circulars, and other regulatory documents. These documents provide authoritative legal sources that can be used to support the answers in the collected consultation data. The resulting corpus also serves as the candidate document collection for the legal information retrieval task.

\subsubsection{Expert Verification of Legal References.}
The answers collected from legal consultation platforms frequently contain explicit references to legal provisions, such as articles, clauses, and sections, cited by legal professionals to support their responses. Rather than treating these references as automatically correct, we employ legal experts to verify whether the cited provisions are valid and relevant to the corresponding questions and answers. For each instance, the experts examine the referenced article, clause, or section, assess whether it provides an appropriate legal basis for answering the question, and identify the corresponding authoritative legal document in the collected corpus. References that are incorrect or irrelevant are not retained as supporting evidence. This verification process produces expert-validated links between real-world legal questions and specific legal provisions, which subsequently serve as relevance annotations for the legal information retrieval task.

\begin{figure}[ht]
    \centering
    \includegraphics[width=\linewidth]{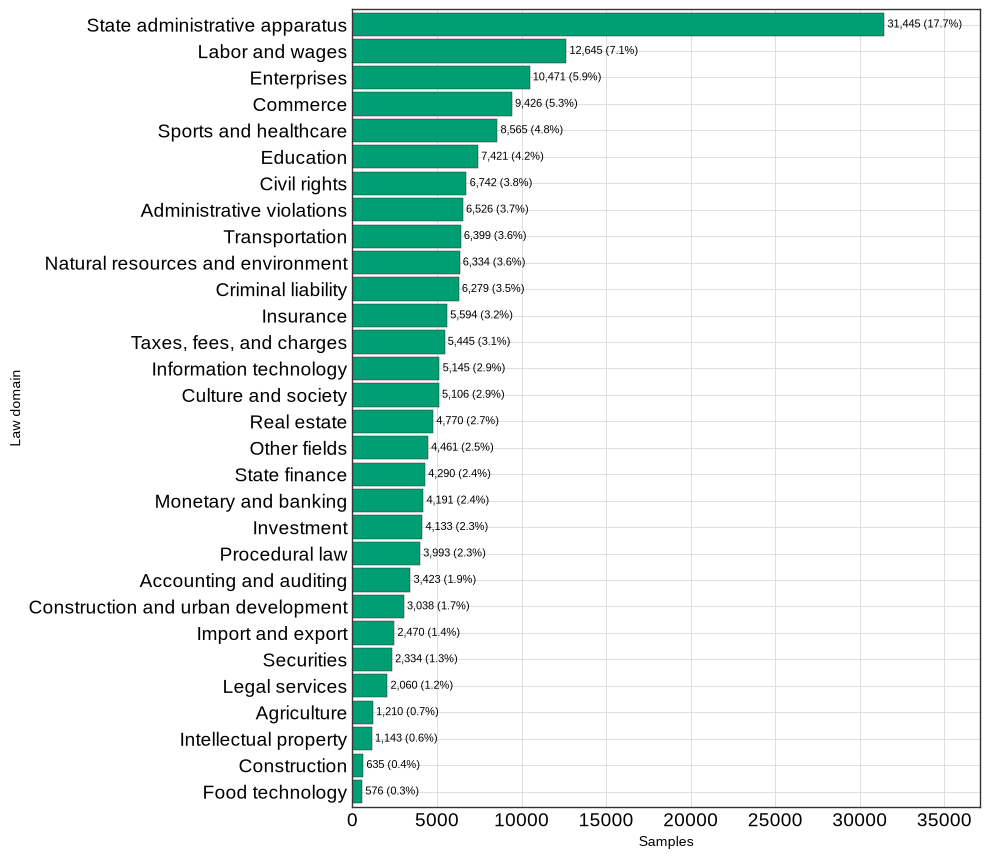}
    \caption{Distribution of legal domains in the proposed dataset.}
    \label{fig:domain_distribution}
\end{figure}

\subsubsection{Data Formatting and Splitting.}

Finally, we normalize and organize the collected data into a unified representation that preserves the original questions and professional answers together with their expert-verified legal references. Each annotated instance therefore connects a naturally occurring legal question to its answer and the corresponding legal provision, including the relevant article, clause, or section and its source legal document. Based on these annotations, we construct the question answering and legal information retrieval subsets and partition the resulting data into training, development, and test sets for standardized evaluation.

\subsection{Dataset Statistics} \label{sec:dataset_statistics}

\begin{table}[ht]
    \centering
    \setlength{\tabcolsep}{5pt}
    \renewcommand{\arraystretch}{1.1}
    \small
    \caption{Overall statistics of ViLegalExpert.}
    \label{tab:dataset_statistics}
    \begin{tabular}{lr}
    \toprule
    \textbf{Statistic} & \textbf{Value} \\
    \midrule
    \#IR Samples & 87,695 \\
    \#QA Samples & 177,158 \\
    \#Legal Domains & 34 \\
    \#Legal Documents & 9,563 \\
    Average Question Length & 21.08 tokens \\
    Median Question Length & 20 tokens \\
    Average Answer Length & 380.51 tokens \\
    Median Answer Length & 343 tokens \\
    \bottomrule
    \end{tabular}
\end{table}

Table~\ref{tab:dataset_statistics} summarizes ViLegalExpert, which contains \textbf{117,158 QA instances} and \textbf{87,695 retrieval instances} derived from over \textbf{172K} authentic consultation questions across \textbf{34 legal domains}. Figure~\ref{fig:domain_distribution} further shows that the dataset covers a broad range of legal topics with varying frequencies, reflecting diverse real-world legal information needs.

\begin{figure}[ht]
    \centering
    \includegraphics[width=0.85\linewidth]{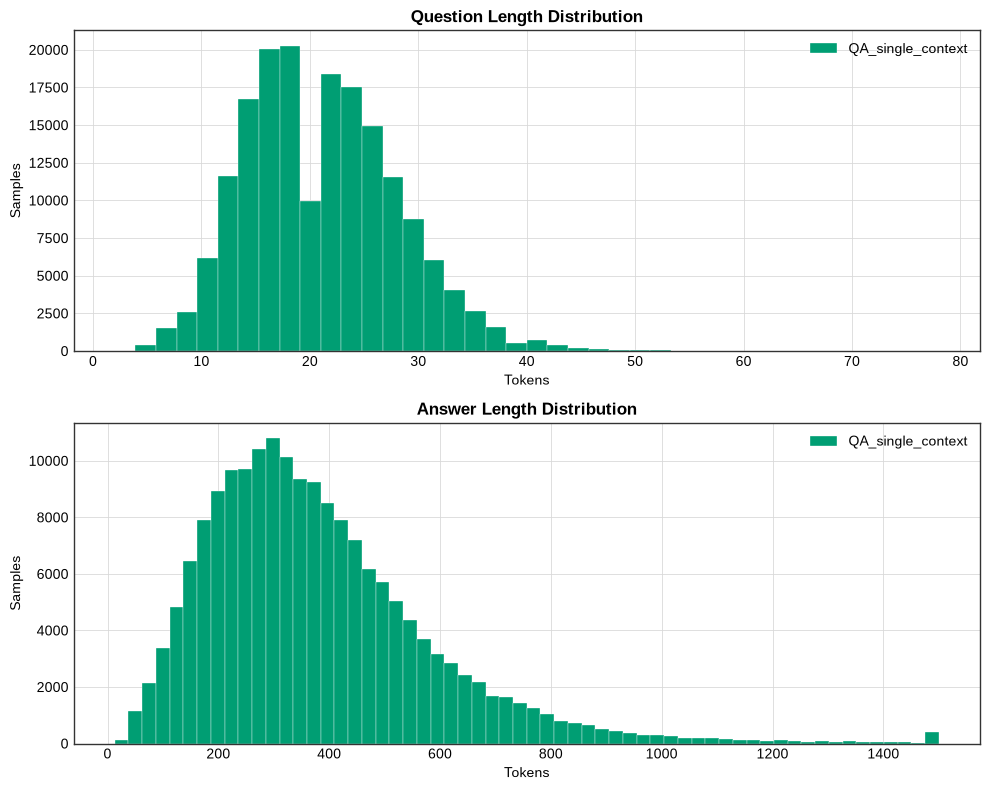}
    \caption{Token-length distributions of questions and answers in ViLegalExpert.}
    \label{fig:answer_question_length}
\end{figure}

\begin{figure}[ht]
    \centering
    \includegraphics[width=0.85\linewidth]{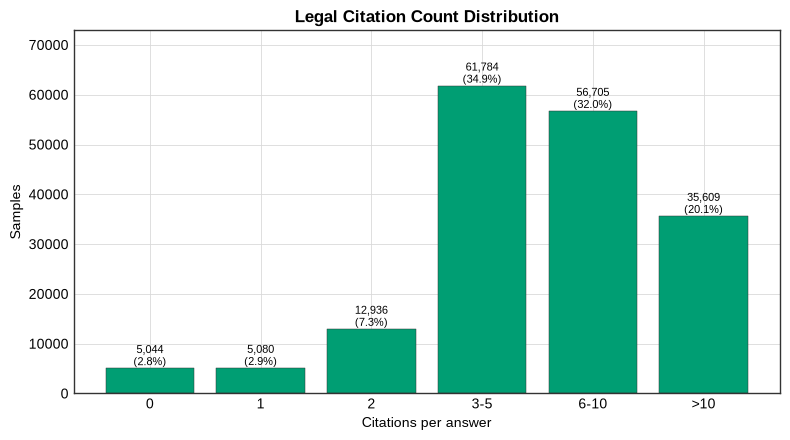}
    \caption{Distribution of legal domains in the proposed dataset.}
    \label{fig:question_variety}
\end{figure}

Figure~\ref{fig:answer_question_length} presents the distributions of question and answer lengths, showing substantial variation in both user queries and professional responses. This reflects the complexity of real-world legal consultations, ranging from concise inquiries to detailed questions and explanations. Figure~\ref{fig:question_variety} further illustrates the diversity of question types in ViLegalExpert, highlighting the heterogeneous legal information needs captured by the dataset.

\section{Experimental Setup} \label{sec:experimental_setup}

We evaluate ViLegalExpert on its three supported tasks: legal information retrieval (IR), extractive question answering (Extractive QA), and abstractive question answering (Abstractive QA). The experiments are designed to assess the benchmark across different modeling paradigms, ranging from conventional retrieval and machine reading comprehension methods to pretrained language models (PLMs) and recent instruction-tuned large language models (LLMs).

\subsection{Experimental Configuration} \label{sec:experimental_configuration}

All experiments are implemented in PyTorch and conducted on a single NVIDIA A100 GPU with 40\,GB of memory. Trainable models are optimized on the training set, with the best checkpoints selected based on development-set performance. We use publicly available checkpoints and their corresponding tokenizers, with model-specific hyperparameters following the recommended configurations.

For QA, models are provided with the question and its associated legal context to either extract an answer span or generate a free-form response. For IR, each question is used as a query over the legal corpus, with expert-verified legal provisions serving as relevance annotations. Instruction-tuned LLMs are evaluated using a consistent prompting template to ensure comparability across models.

\subsection{Baselines and Evaluation Metrics} \label{sec:baselines_metrics}

We evaluate representative baselines for all three tasks supported by ViLegalExpert: legal information retrieval, extractive question answering, and abstractive question answering. The selected methods cover conventional task-specific architectures, pretrained language models, retrieval-augmented approaches, and recent instruction-tuned large language models.

\subsubsection{Legal Information Retrieval.}
We evaluate several representative retrieval approaches, including MiniRAG \cite{fan2025minirag}, LightRAG \cite{guo2024lightrag}, HippoRAG \cite{gutierrez2024hipporag}, and RAPTOR \cite{sarthi2024raptor}. These methods represent recent retrieval and retrieval-augmented architectures that exploit different strategies for organizing and retrieving information from large document collections. We additionally evaluate BM25 \cite{10.1561/1500000019} under sparse, dense, and hybrid retrieval settings \cite{ma2021replicationstudydensepassage}. All methods retrieve from the same legal corpus and are evaluated against the expert-verified legal references in ViLegalExpert.

\subsubsection{Extractive Question Answering.}
For extractive QA, we consider two groups of baselines. The first consists of established machine reading comprehension models, including QANet \cite{yu2018qanet}, DEEP-CASCADE \cite{Yan_2019}, and TD-SAN \cite{zhuang2019token}. The second consists of Transformer-based pretrained language models, including LEGAL-BERT \cite{chalkidis2020legalbertmuppetsstraightlaw}, ViDeBERTa \cite{tran-etal-2023-videberta}, multilingual BERT (mBERT) \cite{pires-etal-2019-multilingual}, and PhoBERT \cite{nguyen2020phobertpretrainedlanguagemodels}. These models enable comparisons among task-specific MRC architectures, legal-domain pretrained representations, multilingual models, and Vietnamese-specific pretrained language models.

\subsubsection{Abstractive Question Answering.}
For abstractive QA, we evaluate models from three representative families. First, we consider conventional MRC and neural generation approaches, including CPG \cite{tay-etal-2019-simple}, S-NET \cite{tan2018snetanswerextractionanswer}, LatentQA \cite{Bi_Wu_Yan_Wang_Xia_Li_2020}, DCMM+ \cite{zhang2020dcmndualcomatchingnetwork}, MultiStyle \cite{nishida2019multistylegenerativereadingcomprehension}, GAQA \cite{roy-etal-2022-investigating}, and CHIME \cite{lu-etal-2020-chime}. Second, we evaluate pretrained sequence-to-sequence language models, including ViT5 \cite{phan-etal-2022-vit5}, BARTpho \cite{tran2022bartphopretrainedsequencetosequencemodels}, mT5 \cite{xue2021mt5massivelymultilingualpretrained}, and mBART \cite{liu2020multilingualdenoisingpretrainingneural}. Finally, we evaluate recent open-source instruction-tuned LLMs, including Qwen2.5-7B-Instruct, Llama3.1-8B-Instruct, Llama3-8B-Instruct, Qwen2.5-14B-Instruct, and Mistral-7B-Instruct-v0.3. This diverse set of baselines allows us to assess the progression from specialized neural architectures to pretrained sequence-to-sequence models and modern instruction-tuned LLMs on real-world Vietnamese legal questions.

\subsubsection{Evaluation Metrics.}
For both extractive and abstractive QA, we report \textbf{ROUGE-L}, \textbf{METEOR}, and \textbf{BERTScore} to measure lexical overlap and semantic similarity between predicted and reference answers. For legal IR, we use \textbf{F1}, \textbf{Precision}, and \textbf{Mean Reciprocal Rank (MRR)} to evaluate retrieval accuracy and ranking quality. Higher values indicate better performance for all metrics.

\section{Results} \label{sec:results}


\subsection{Question Answering Results}

\begin{table}[ht]
    \centering
    \setlength{\tabcolsep}{4pt}
    \renewcommand{\arraystretch}{1.02}
    \caption{Extractive QA results. R-L, MET, and BS denote ROUGE-L, METEOR, and BERTScore, respectively. Best results within each model group are in bold.}
    \label{tab:extractive_qa}
    \begin{tabular}{lccc}
    \toprule
    \textbf{Method} & \textbf{R-L} & \textbf{MET} & \textbf{BS} \\
    \midrule
    \multicolumn{4}{c}{\textit{MRC Models}} \\
    \midrule
    QANet \cite{yu2018qanet} & \textbf{35.06} & \textbf{40.12} & 61.06 \\
    DEEP-CASCADE \cite{Yan_2019} & 34.80 & 40.10 & \textbf{72.09} \\
    TD-SAN \cite{zhuang2019token} & 31.34 & 36.63 & 70.37 \\
    \midrule
    \multicolumn{4}{c}{\textit{Pre-trained Language Models}} \\
    \midrule
    LEGAL-BERT \cite{chalkidis2020legalbertmuppetsstraightlaw} & 62.48 & 64.60 & 84.08 \\
    ViDeBERTa \cite{tran-etal-2023-videberta} & \textbf{62.79} & \textbf{66.08} & \textbf{86.27} \\
    mBERT \cite{pires-etal-2019-multilingual} & 62.36 & 64.26 & 86.17 \\
    PhoBERT \cite{nguyen2020phobertpretrainedlanguagemodels} & 50.68 & 53.54 & 81.84 \\
    \bottomrule
    \end{tabular}
\end{table}

\subsubsection{Extractive Question Answering.} Table~\ref{tab:extractive_qa} shows a substantial performance gap between conventional MRC architectures and pretrained language models. Among the MRC models, QANet achieves the highest ROUGE-L (35.06) and METEOR (40.12), while DEEP-CASCADE obtains the highest BERTScore of 72.09. In contrast, all evaluated pretrained language models achieve considerably stronger results, demonstrating the effectiveness of contextualized pretrained representations for Vietnamese legal answer extraction.

Among the pretrained models, ViDeBERTa performs best across all three metrics, achieving 62.79 ROUGE-L, 66.08 METEOR, and 86.27 BERTScore. mBERT closely follows in ROUGE-L and BERTScore, whereas LEGAL-BERT also demonstrates competitive performance despite not being specifically pretrained for Vietnamese. Interestingly, PhoBERT performs below the other pretrained models, suggesting that Vietnamese-specific pretraining alone does not necessarily provide an advantage for legal answer extraction. Overall, the results indicate that pretrained contextual representations provide a substantial improvement over conventional MRC architectures, while leaving further room for domain- and task-specific adaptation.

\begin{table}[ht]
    \centering
    \setlength{\tabcolsep}{4pt}
    \renewcommand{\arraystretch}{1.02}
    \caption{Abstractive QA results. R-L, MET, and BS denote ROUGE-L, METEOR, and BERTScore, respectively. Best results within each model group are in bold.}
    \label{tab:abstractive_qa}
    \begin{tabular}{lccc}
    \toprule
    \textbf{Method} & \textbf{R-L} & \textbf{MET} & \textbf{BS} \\
    \midrule
    \multicolumn{4}{c}{\textit{MRC Models}} \\
    \midrule
    CPG \cite{tay-etal-2019-simple} & \textbf{29.85} & \textbf{26.15} & \textbf{73.22} \\
    S-NET \cite{tan2018snetanswerextractionanswer} & 26.39 & 24.02 & 70.42 \\
    LatentQA \cite{Bi_Wu_Yan_Wang_Xia_Li_2020} & 26.45 & 24.30 & 71.26 \\
    DCMM+ \cite{zhang2020dcmndualcomatchingnetwork} & 25.84 & 23.75 & 70.73 \\
    MultiStyle \cite{nishida2019multistylegenerativereadingcomprehension} & 25.06 & 23.11 & 70.35 \\
    GAQA \cite{roy-etal-2022-investigating} & 25.93 & 23.82 & 70.85 \\
    CHIME \cite{lu-etal-2020-chime} & 25.59 & 23.55 & 70.84 \\
    \midrule
    \multicolumn{4}{c}{\textit{Pre-trained Language Models}} \\
    \midrule
    ViT5 \cite{phan-etal-2022-vit5} & 40.59 & 44.74 & 76.37 \\
    BARTpho \cite{tran2022bartphopretrainedsequencetosequencemodels} & \textbf{44.36} & \textbf{49.40} & \textbf{78.52} \\
    mT5 \cite{xue2021mt5massivelymultilingualpretrained} & 30.55 & 30.17 & 72.82 \\
    mBART \cite{mbart} & 39.34 & 45.26 & 76.20 \\
    \midrule
    \multicolumn{4}{c}{\textit{Open-source Language Models}} \\
    \midrule
    Qwen2.5-7B-Instruct & 24.20 & 22.57 & 71.34 \\
    Llama3.1-8B-Instruct & 25.15 & 23.20 & 72.10 \\
    Llama3-8B-Instruct & \textbf{28.31} & \textbf{28.69} & \textbf{73.00} \\
    Qwen2.5-14B-Instruct & 23.04 & 21.44 & 70.12 \\
    Mistral-7B-Instruct-v0.3 & 12.40 & 12.94 & 63.98 \\
    \bottomrule
    \end{tabular}
\end{table}

\subsubsection{Abstractive Question Answering.} Table~\ref{tab:abstractive_qa} presents a different performance pattern. Among conventional MRC and generation models, CPG achieves the strongest overall results, with 29.85 ROUGE-L, 26.15 METEOR, and 73.22 BERTScore. Nevertheless, these models are substantially outperformed by pretrained sequence-to-sequence language models. BARTpho achieves the best performance across all evaluated approaches, reaching 44.36 ROUGE-L, 49.40 METEOR, and 78.52 BERTScore. ViT5 also performs strongly, particularly in METEOR and BERTScore.

Surprisingly, the evaluated instruction-tuned LLMs do not outperform the smaller pretrained sequence-to-sequence models. Llama3-8B-Instruct is the strongest model in this group, with 28.31 ROUGE-L, 28.69 METEOR, and 73.00 BERTScore, but remains substantially behind BARTpho. Increasing model size also does not consistently improve performance: Qwen2.5-14B-Instruct performs below its 7B counterpart across all three metrics. These results suggest that general instruction-following capability and model scale alone are insufficient for Vietnamese legal question answering. In contrast, language-specific sequence-to-sequence pretraining appears particularly effective when models are adapted to the target task.

\subsection{Legal Information Retrieval Results}

\begin{table}[ht]
    \centering
    \setlength{\tabcolsep}{4pt}
    \renewcommand{\arraystretch}{1.02}
    \caption{Legal information retrieval results. P denotes Precision. Best overall results are in bold.}
    \label{tab:retrieval_results}
    \begin{tabular}{lccc}
    \toprule
    \textbf{Method} & \textbf{F1} & \textbf{MRR} & \textbf{P} \\
    \midrule
    MiniRAG \cite{fan2025minirag} & 12.28 & 17.73 & 8.32 \\
    LightRAG \cite{guo2024lightrag} & 15.96 & 20.53 & 9.58 \\
    HippoRAG \cite{gutierrez2024hipporag} & 15.64 & 12.63 & 6.40 \\
    RAPTOR \cite{sarthi2024raptor} & 14.80 & 21.44 & 10.03 \\
    \midrule
    BM25 (Sparse) \cite{10.1561/1500000019} & 16.03 & 23.23 & 10.86 \\
    BM25 (Dense) \cite{ma2021replicationstudydensepassage} & 12.40 & 17.51 & 8.41 \\
    BM25 (Hybrid) & \textbf{18.06} & \textbf{26.16} & \textbf{12.24} \\
    \bottomrule
    \end{tabular}
\end{table}

Table~\ref{tab:retrieval_results} shows that retrieving supporting legal evidence for real-world consultation questions remains challenging. Among the recent retrieval and RAG-based approaches, LightRAG achieves the highest F1 score (15.96), while RAPTOR obtains the highest MRR (21.44) and Precision (10.03). HippoRAG performs comparably in F1 but exhibits a considerably lower MRR and Precision, indicating differences in the ability of these methods to rank relevant legal evidence near the top of the retrieved results.

The sparse BM25 setting provides a strong baseline, achieving 16.03 F1, 23.23 MRR, and 10.86 Precision and outperforming all evaluated RAG-based retrieval methods on these metrics. This result suggests that lexical matching remains particularly effective for Vietnamese legal retrieval. Legal texts contain highly standardized terminology and recurring lexical expressions, providing useful exact-match signals that can be difficult to preserve using semantic representations alone.

The hybrid retrieval setting achieves the best overall performance, with 18.06 F1, 26.16 MRR, and 12.24 Precision. Its consistent improvement over both sparse and dense settings indicates that lexical and semantic signals are complementary: sparse retrieval effectively captures terminology-level correspondence, whereas dense representations can identify semantically related evidence when exact lexical overlap is insufficient. Nevertheless, the relatively low absolute performance of all retrieval approaches highlights the difficulty of mapping naturally expressed citizen questions to the appropriate provisions in formal legal documents.

Overall, across the three tasks, the results reveal different strengths of current modeling paradigms. Pretrained encoder models substantially outperform conventional MRC architectures for extractive QA, while Vietnamese-oriented sequence-to-sequence models are particularly effective for abstractive QA and outperform the evaluated instruction-tuned LLMs. For legal information retrieval, lexical retrieval remains a strong baseline, with the hybrid setting providing further improvements by combining lexical and semantic signals. At the same time, none of the evaluated approaches achieves consistently strong performance across all tasks, demonstrating the challenges posed by real-world Vietnamese legal consultation data and leaving substantial room for future advances.

\section{Conclusion} \label{sec:conclusion}

We introduce \textbf{ViLegalExpert}, a large-scale Vietnamese benchmark for legal retrieval and question answering constructed from authentic citizen--lawyer consultations with expert-verified legal evidence. Covering retrieval, extractive QA, and abstractive QA, our experiments reveal substantial challenges in mapping real-world legal questions to authoritative provisions and generating correctly grounded answers. We hope ViLegalExpert will support future research toward more reliable and evidence-grounded Vietnamese Legal AI.


\bibliographystyle{splncs04}
\bibliography{main}

\appendix

\end{document}